\pdfoutput=1

\documentclass[11pt,a4paper]{article}
\usepackage[hyperref]{emnlp2020}
\usepackage{times}
\usepackage{latexsym}

\usepackage{microtype}

\aclfinalcopy 

\usepackage{graphicx}
\usepackage{eurosym}
\usepackage{tablefootnote}
\usepackage{kantlipsum}
\graphicspath{ {images/} }
\usepackage{hyperref}
\usepackage{textcomp}

\usepackage{hhline}
\usepackage[utf8]{inputenc}

\usepackage[T1]{fontenc}
\usepackage{listings}
\usepackage{stfloats}
\usepackage{csquotes}
\usepackage{array}
\usepackage{booktabs}

\usepackage[super]{nth}
\usepackage[inline,shortlabels]{enumitem}
\usepackage{multirow}
\usepackage{makecell}
\usepackage{siunitx}
\usepackage{multicol}

\newcounter{notecounter}
\newcommand{\enotesoff}{\long\gdef\enote##1##2{}}
\newcommand{\enoteson}{\long\gdef\enote##1##2{{
			\stepcounter{notecounter}
			\large\bf
			\hspace{100cm}\arabic{notecounter} $<<<$ ##1: ##2
			$>>>$\hspace{1cm}}}}
\enoteson
\enotesoff

\usepackage{svg}
\usepackage{float}
\usepackage{ amssymb }
\usepackage{amsmath,amsfonts,amssymb}

\title{Reusing a Pretrained Language Model \\ on Languages with Limited Corpora for Unsupervised NMT}

\author{ 
	Alexandra Chronopoulou,
    Dario Stojanovski,
	Alexander Fraser\\\\
	Center for Information and Language Processing, LMU Munich, Germany \\
	{\tt \{achron, stojanovski, fraser\}@cis.lmu.de}
	}
	
\date{}

\begin{document}
\maketitle

\begin{abstract}

Using a language model (\textsc{lm}) pretrained on two languages with large monolingual data in order to initialize an unsupervised neural machine translation (\textsc{unmt}) system 
yields state-of-the-art results. 
When limited data is available for one language, however, this method leads to poor translations. We present an effective approach that reuses an \textsc{lm} that is pretrained
only on a high-resource language. 
The monolingual \textsc{lm} is fine-tuned on both languages and is then used to initialize a \textsc{unmt} model.
To reuse the pretrained \textsc{lm}, we have to modify its predefined vocabulary, to account for the new language. We therefore propose a novel vocabulary extension method. 
Our approach, \textsc{re-lm}, outperforms a competitive cross-lingual pretraining model (\textsc{xlm}) in English-Macedonian (En-Mk)
and English-Albanian (En-Sq), yielding more than $+8.3$ \textsc{bleu} points for all four translation directions.

\end{abstract}
\section{Introduction}

Neural machine translation (\textsc{nmt}) has recently achieved remarkable results  \cite{bahdanau2015, vaswani2017attention}, based on the exploitation of large parallel training corpora. Such corpora are 
only available for a limited number of
languages. \textsc{unmt} has attempted to address this limitation by training \textsc{nmt} systems using  monolingual data \textit{only} \cite{artetxe2017unsupervised, lample2017unsupervised}.
Top performance is achieved using a bilingual masked language model \cite{devlin-etal-2019-bert} to initialize a \textsc{unmt} encoder-decoder system \cite{lample2019cross}. The model is then trained using denoising auto-encoding \cite{vincent2008extracting} and back-translation \cite{sennrich2015improving}. The approach was mainly evaluated by translating between high-resource languages.

Translating between a high-resource and a low-resource language is a more challenging task.
In this setting, the \textsc{unmt} model can be initialized with a pretrained cross-lingual \textsc{lm}. However, training this \textsc{unmt} model has been
shown to be ineffective when the two languages are not related \cite{guzman2019flores}.
Moreover, in order to use a pretrained cross-lingual \textsc{lm} to initialize a \textsc{unmt} model, the two models must have a shared vocabulary. Thus, a bilingual \textsc{lm} needs to be trained from scratch for each language pair, before being transferred to the \textsc{unmt} model
(e.g. En-De \textsc{lm} for En-De \textsc{unmt}).

Motivated by these issues, we focus on the question: 
\textit{how can we accurately and efficiently translate between a \textit{high-monolingual-resource} (\textbf{\textsc{hmr}}) and a \textit{low-monolingual-resource} (\textbf{\textsc{lmr}}) language}?    
To address this question, we adapt a monolingual \textsc{lm}, pretrained on an \textsc{hmr} language to an \textsc{lmr} language, in order to initialize a  \textsc{unmt} system.

We make the following contributions:
(1) We propose \textsc{re}used-\textsc{lm}\footnote{We release the code in \url{https://github.com/alexandra-chron/relm_unmt}.} (\textbf{\textsc{re-lm}}),
an effective transfer learning method for \textsc{unmt}. Our method
reuses a pretrained \textsc{lm} on an \textsc{hmr} language, by fine-tuning it on both \textsc{lmr} and \textsc{hmr} languages. The fine-tuned \textsc{lm} is used to initialize
a \textsc{unmt} system that translates the \textsc{lmr} to the \textsc{hmr} language (and vice versa). 
(2) We introduce a novel vocabulary extension method, which allows fine-tuning a pretrained \textsc{lm} to an unseen language. 
(3) We show that \textsc{re-lm} outperforms a competitive transfer learning method (\textsc{xlm})  for \textsc{unmt} on three language pairs: English-German (En-De) on a synthetic setup, En-Mk and En-Sq. 
(4) We show that \textsc{re-lm} is effective in low-resource supervised \textsc{nmt}. 
(5) We conduct an analysis of fine-tuning schemes for \textsc{re-lm} 
and find that including adapters \cite{Houlsby2019ParameterEfficientTL} in the training procedure yields almost the same  \textsc{unmt} results as \textsc{re-lm} at a lower computational price. We also run experiments to identify the contribution of the vocabulary extension method. 
\enote{DS}{Are we going to call RE-LM only the fine-tuning on both approach? It is cleaner that way. Not sure if it would make sense to encompass all methods that use vocab ext as RE-LM, because all use reusing. Or just give some name to RE-LM with adapters.} \enote{AC}{I think it's cleaner this way. I was considering calling the + adapters ft on LMR -> adapter RE-LM. Not sure though.}
\enote{DS}{That seems OK to me.}

\vspace{-4pt}
\section{Related Work}

\begin{figure}[]
	\centering
	\includegraphics[width=0.9\columnwidth, page=1]{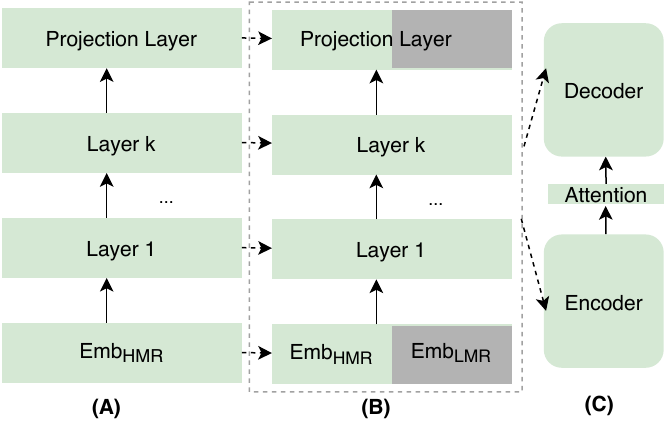}
	\caption{\textbf{\textsc{re-lm}}. \textbf{(A)} \textsc{lm} pretraining. \textbf{(B)} Fine-tuning. The embedding and the projection layer are extended using  \S \ref{ssec:voctrick} (dark gray) and  \textbf{(C)} Transfer to an \textsc{nmt} system. Dashed arrows indicate transfer of weights. }\label{fig:lru}
\end{figure}

\noindent \textbf{Transfer learning for \textsc{unmt}.} 
The field of \textsc{unmt} has recently experienced tremendous progress.
\citet{artetxe2017unsupervised, lample2017unsupervised} train \textsc{unmt} models with  monolingual data only, using denoising auto-encoding \cite{vincent2008extracting} and online back-translation \cite{sennrich2015improving} as training objectives.
 This approach is successful for languages with high-quality, large, comparable data. When these conditions are not met, though, \textsc{unmt} provides near-zero scores \cite{neubig-hu-2018-rapid}. 
\textsc{unmt} is further improved when initialized with a cross-lingual pretrained model, trained on large corpora \cite{lample2019cross,song2019mass}. However, many languages have only limited monolingual data available, a setting where \textsc{unmt} is not effective \cite{guzman2019flores}. \citet{sun2020self}, whose work is close to our work in motivation, train a \textsc{unmt} model for an \textsc{hmr}-\textsc{lmr} language pair. Iteratively, every subset (e.g. $10$\%) of \textsc{hmr} and all \textsc{lmr} data is backtranslated and the pseudo-parallel corpus is added to the training process. Just like \textsc{xlm}, this training procedure needs to run from scratch for every new language pair. By contrast, our method fine-tunes a monolingual pretrained \textsc{lm} for \textsc{unmt}, so it is computationally faster and simpler. 


\noindent \textbf{Vocabulary.} Transferring a pretrained model (\textit{source}) to a new model (\textit{target}) requires the use of a shared vocabulary \cite{nguyen-chiang-2017-transfer}. 
\citet{kim-etal-2019-effective} propose a linear alignment of the source and target model embeddings using an unsupervised dictionary. 
However, when the embeddings of the two models do not have enough overlapping strings, dictionary induction might fail \cite{sogaard-etal-2018}. \citet{dynamicvoc} transfer a source \textsc{nmt} model to a target \textsc{nmt} model (e.g. De-En to Nl-En). To enable transfer, they overwrite the source vocabulary with the target vocabulary. By contrast, we keep the union of the two vocabularies. We fine-tune a pretrained monolingual \textsc{lm} to an \textsc{lmr} language, to initialize an \textsc{nmt} model. Thus, we need the vocabularies of both languages.

\noindent \textbf{Adapters.} 
Residual adapters \cite{Houlsby2019ParameterEfficientTL} are feed-forward networks, added to each of to the original model's layers.  During fine-tuning, the model parameters are frozen and only the adapters are fine-tuned. This can prevent catastrophic forgetting \cite{Goodfellow2013AnEI,bapna-firat-2019-simple}.
Adapters show promising results in domain adaptation \cite{bapna-firat-2019-simple} and cross-lingual classification \cite{artetxe2019cross}.
Motivated by this, we study the use of adapters during \textsc{lm} fine-tuning in our analysis.

\section{Proposed Approach}
\label{sec:approach}

We describe our method
for translation between a high-resource (\textsc{hmr}) and a low-resource language (\textsc{lmr})
using
monolingual data in this section. 
\vspace{-3pt}
\subsection{\textsc{re-lm}}
\label{ssec:remlm}
Our proposed approach consists of three steps, as shown in Figure~\ref{fig:lru}: 

\noindent 
\textbf{(A)} We train a monolingual masked \textsc{lm} on the \textsc{hmr} language, using all available \textsc{hmr} corpora. This step needs to be performed only \textit{once} for the \textsc{hmr} language.
Note that a publicly available pretrained model could also be used.

\noindent 
\textbf{(B)} To fine-tune the pretrained \textsc{lm} on the \textsc{lmr} language, we first need to overcome the vocabulary mismatch problem. 
Fine-tuning without extending the vocabulary is detrimental, as we will show later in the analysis. We therefore extend the vocabulary of the pretrained model using our proposed method, described in \S \ref{ssec:voctrick}. 

\noindent \textbf{(C)} Finally, we initialize an encoder-decoder \textsc{unmt} system with the fine-tuned \textsc{lm}. The \textsc{unmt} model is trained using denoising auto-encoding and online back-translation for the \textsc{hmr-lmr} language pair. 
 \begin{figure}[h]
	\centering
	\includegraphics[width=0.93\columnwidth, page=1]{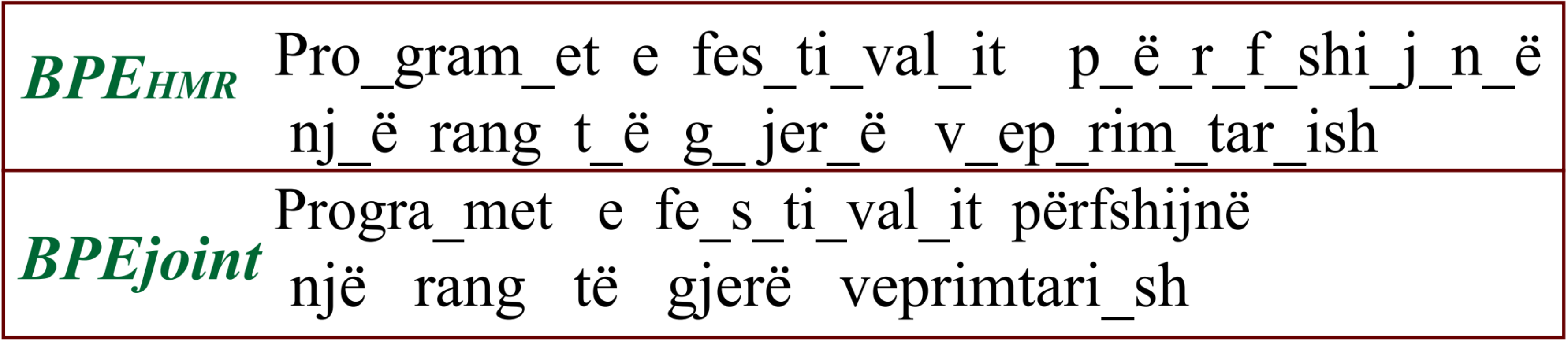}
	\caption{Segmentations of Albanian (Sq). We observe that splitting Sq using En BPEs ($\text{BPE}_{\text{\textit{HMR}}}$) results in heavily segmented tokens. This problem is alleviated using $\text{BPE}_{joint}$ tokens, learned on both languages.}
	\label{fig:german}
\end{figure}
\setlength{\belowcaptionskip}{-10pt}
\subsection{Vocabulary Extension}
\label{ssec:voctrick}

We propose a novel method that enables adapting a pretrained monolingual \textsc{lm} to an unseen language. We consider the case of an \textsc{lm} pretrained on an \textsc{hmr} language. The training data is split using Byte-Pair-Encoding (BPE) \cite{sennrich2015neural}. We denote these BPE tokens as $\text{BPE}_{\text{\textit{HMR}}}$ and the resulting vocabulary as $V_{\text{\textit{HMR}}}$. We aim to fine-tune the trained \textsc{lm} to an unseen \textsc{lmr} language. 
Splitting the \textsc{lmr} language with $\text{BPE}_{\text{\textit{HMR}}}$ tokens would result in heavy segmentation of \textsc{lmr} words (Figure \ref{fig:german}). To counter this, we learn BPEs on the joint \textsc{lmr} and \textsc{hmr} corpus ($\text{BPE}_{joint}$).
We then use $\text{BPE}_{joint}$ tokens to split the \textsc{lmr} data, resulting in a vocabulary $V_{\text{\textit{LMR}}}$. This technique increases the number of shared tokens and enables  cross-lingual transfer of the pretrained \textsc{lm}. The final vocabulary is the union of the $V_{\text{\textit{HMR}}}$ and $V_{\text{\textit{LMR}}}$ vocabularies. 
We extend the input and output embedding layer to account for the new vocabulary items. The new parameters are then learned during fine-tuning. 

\section{Experimental Setup}

\noindent \textbf{Datasets.} We experiment with two setups.
In the first \textit{synthetic} setup we use En-De. We sample $8$M En sentences from NewsCrawl. To simulate an \textsc{lmr} language, we gradually sample $0.05$M, $0.5$M and $1$M De sentences. 
We use the WMT dev/test sets \cite{bojar-etal-2016-findings}. The second, \textit{real-world setup} is En-Mk, En-Sq. We use $68$M En sentences from NewsCrawl. For Mk and Sq, we use $2.4$M Mk and $4$M Sq, obtained from OSCAR\footnote{\url{https://oscar-corpus.com/}} \cite{ortizsuarez:hal-02148693} and Wikipedia. We randomly select $3$K sentences from SETIMES\footnote{\url{http://opus.nlpl.eu/SETIMES.php}} as dev and $3$K as test set.
We tokenize data with standard Moses \cite{koehn2006open} scripts.
For the low-resource supervised case, we sample $10$K, $100$K, and $200$K parallel sentences from SETIMES for Mk and Sq.

\noindent \textbf{Preprocessing}. 
We train a standard \textsc{xlm} model \cite{lample2019cross} as a baseline using $32$K BPE merge operations, learned on the concatenation of sentences sampled randomly from the corpora of each language pair with $\alpha = 0.5$. For \textsc{re-lm}, we learn $32$K BPEs on the \textsc{hmr} corpus and extract the initial vocabulary ($V_{\text{\textit{HMR}}}$). Then, we learn $32$K BPEs on the joint \textsc{lmr} and \textsc{hmr} corpus ($\text{BPE}_{joint}$). We 
extend the initial $\text{V}_{\text{\textit{HMR}}}$ vocabulary by the amount of   \textsc{lmr} vocabulary items that are not already present in $\text{V}_{\text{\textit{HMR}}}$. To identify whether a smaller number of BPE merges would be useful for splitting the \textsc{lmr} language, we conduct experiments varying their number in the analysis.

\noindent \textbf{Model Configuration.} \textsc{re-lm} is built using the \textsc{xlm} codebase\footnote{\href{https://github.com/facebookresearch/XLM/}{\texttt{github.com/facebookresearch/XLM/}}}. Each masked \textsc{lm} has a Transformer architecture with $1024$ hidden units, $6$ layers and $8$ attention heads. Each \textsc{nmt} model is a $6$-layer encoder-decoder Transformer with $1024$ hidden units and $8$ heads. 
Each \textsc{lm} is trained using Adam \cite{kingma2014adam} with learning rate $10^{-4}$ and masking follows \citet{devlin-etal-2019-bert}. During \textsc{unmt} and supervised \textsc{nmt} training, Adam with inverse square root scheduling and a learning rate of $10^{-4}$ is used.  We evaluate \textsc{nmt} models on the dev set every $3000$ updates using greedy decoding. The En \textsc{lm} and each \textsc{xlm} are trained on $8$ NVIDIA GTX $11$ GB GPUs for $1$ week, with a per-GPU batch size of $32$. \textsc{lm} fine-tuning and \textsc{nmt} training models are computationally efficient, using just $1$ GPU and $32$ batch size. We assume that by fine-tuning the \textsc{lm} on $8$ GPUs, we could get even better results. 
Final translations are generated using beam search of size $5$. 
We report de-tokenized \textsc{bleu} using Sacre\textsc{bleu} \cite{post-2018-call}\footnote{Signature ``BLEU+c.mixed+\#.1+s.exp+tok.13a+v.1.4.9''}.

\noindent \textbf{Experiments.} For unsupervised translation, we train a randomly initialized \textsc{unmt} model for each language pair as a first baseline. As a transfer learning baseline, we use \textsc{xlm} \cite{lample2019cross}, trained on the two languages and transferred to a \textsc{unmt} model. The \textsc{unmt} models are trained using monolingual data. 
For supervised translation, \textsc{nmt} training is performed using only parallel corpora, without offline back-translation of monolingual data. The first baseline is a randomly initialized \textsc{nmt} system. The second baseline is an \textsc{nmt} model initialized with \textsc{xlm}. We compare them to our proposed approach, \textsc{re-lm}.
Both \textsc{xlm} and \textsc{re-lm} are trained on the monolingual corpora of both languages of interest. 
In the analysis, we add adapters \cite{rebuffi18} of hidden size $256$ after each self-attention and each feed-forward layer of the pretrained monolingual \textsc{lm}. We freeze the parameters of the pretrained \textsc{lm} and fine-tune only the adapters and the embedding layer. 

\section{Results and Analysis}
\subsection{Unsupervised Translation}
Table \ref{table:results} presents our \textsc{unmt} results, comparing random initialization, \textsc{xlm} and \textsc{re-lm}.
\begin{table*}[ht]
\centering
\small

\begin{tabular}{lrrrrrr|rrrr}
\toprule
   \textsc{hmr}-\textsc{lmr} language pair  &  \multicolumn{2}{c}{\textbf{En-De}}  & \multicolumn{2}{c}{\textbf{En-De}}   & \multicolumn{2}{c|}{\textbf{\textbf{En-De}}} & \multicolumn{2}{c}{\textbf{En-Mk}}           & \multicolumn{2}{c}{\textbf{En-Sq}}     \\
   size of \textsc{lmr} language   &  \multicolumn{2}{c}{\textbf{0.05M}}  & \multicolumn{2}{c}{\textbf{0.5M}}   & \multicolumn{2}{c|}{\textbf{1M}} & \multicolumn{2}{c}{\textbf{2.4M}}           & \multicolumn{2}{c}{\textbf{4M}}     \\
    &    $\leftarrow$  &  $\rightarrow$      &   $\leftarrow$  &  $\rightarrow$    &   $\leftarrow$  &  $\rightarrow$       &    $\leftarrow$  &  $\rightarrow$   &   $\leftarrow$  &  $\rightarrow$    \\ 
 \midrule
random       &   3.9 & 4.9   &   3.4  &    2.6    & 4.2 &  4.1        &  3.5    & 3.0 & 6.6  & 5.6      \\
\textsc{xlm}      & 8.1  & 6.4   & 19.8   & 16.0 &  21.7&   18.1          & 12.2 &  12.8  & 16.3   & 18.8   \\
\textsc{re-lm}  & \textbf{10.7} & \textbf{7.5}  &\textbf{22.6}   & \textbf{19.0}   & \textbf{24.3} &\textbf{21.9}  & \textbf{22.0}   & \textbf{21.1}    &  \textbf{27.6}    & \textbf{28.1} \\
\bottomrule

\end{tabular}
\caption{\textsc{unmt} \textsc{bleu} scores. The first column indicates the pretraining method used. Left arrow ($\leftarrow$) refers to translation from the \textsc{lmr} language to En, while right arrow ($\rightarrow$) refers to translation from En to the \textsc{lmr} language.}
\label{table:results}
\end{table*}

\noindent \textbf{Synthetic setup.} We observe that \textsc{re-lm} consistently outperforms \textsc{xlm}. Using $50$K De sentences, 
\textsc{re-lm} has small gains over \textsc{xlm} ($+1.1$ \textsc{bleu} in En$\rightarrow$De). However, when we scale to slightly more data ($500$K), the performance of \textsc{re-lm} is clearly better than the one of \textsc{xlm}, with $+3$ En$\rightarrow$De \textsc{bleu} gains. With $1$M De data, our model surpasses the \textsc{xlm} by more than $2.6$ \textsc{bleu} in both directions.

\noindent \textbf{Real-world setup.}
Our approach surpasses \textsc{xlm} in both language pairs. 
We observe that \textsc{re-lm} achieves at least $+8.3$ \textsc{bleu} over \textsc{xlm} for En-Mk. Our model was first pretrained on En and then fine-tuned on both En and Mk. Therefore, it has  processed \textit{all} En and Mk sentences, obtaining a good cross-lingual representation. 
However, \textsc{xlm} is jointly trained on En and Mk. As a result, it overfits Mk before processing all En data.  
\textsc{re-lm} is similarly effective for En-Sq, achieving an improvement of at least $+9.3$ \textsc{bleu} over \textsc{xlm}. 

\noindent \textbf{Synthetic vs Real-world setup.} The effectiveness of \textsc{re-lm} is pronounced in the real-world setup. We identify two potential reasons. 
First, for En-De, $8$M En is used for \textsc{lm} pretraining, while for En-Mk and En-Sq, $68$M En is used. When \textsc{xlm} is trained on imbalanced \textsc{hmr-lmr} data, it overfits the \textsc{lmr} language. This is more evident for the En-Mk (or En-Sq) than for the En-De \textsc{xlm}, perhaps due to the larger data imbalance. Second, in En-De, we use high-quality corpora for both languages (NewsCrawl), whereas Mk and Sq are trained on low-quality CommonCrawl data. The fact that \textsc{re-mlm} outperforms \textsc{xlm} for Mk and Sq shows that it is more robust to noisy data than the \textsc{xlm}.
\enote{DS}{Not sure about the noisy data argument now, the languages being different may be the dominant factor for these results}

\subsection{Low-Resource Supervised Translation} 
We sample $10$K, $100$K and $200$K of En-Mk and En-Sq bi-text and train supervised \textsc{nmt} systems. 
We compare \textsc{xlm, re-lm} and \textit{random}, an \textsc{nmt} model trained from scratch. We observe (Table \ref{table:parallel}) that \textsc{re-lm} consistently outperforms the baselines when trained on $100$K or less for En-Mk and En-Sq. Using $200$K, though, \textsc{re-lm} yields the same results as \textsc{xlm}. We hypothesize that this happens because SETIMES is a homogeneous domain. Thus, training an \textsc{nmt} model with $200$K is sufficient for competitive results, so both pretraining models provide similar improvements over \textit{random}.

\begin{table}[t]
\centering
\small
\begin{tabular}{clcccc}
\toprule
\multirow{2}{*}{parallel} &languages  & \multicolumn{2}{c}{En-Mk} & \multicolumn{2}{c}{En-Sq}    \\ 
& direction  &    $\leftarrow$  &   $\rightarrow$     &    $\leftarrow$  &   $\rightarrow$    \\ \midrule

\multirow{3}{*}{10K} &random   & 23.4      &23.7   & 25.5  & 18.9  \\
&\textsc{xlm}                              & 38.7      & 38.7  & 44.7  & 41.4 \\       
&\textsc{re-lm}                           & \textbf{40.1}      & 38.9  & \textbf{45.7} & \textbf{42.8}      \\ \midrule

\multirow{3}{*}{100K} &random   & 48.4     & 48.2   & 51.8  & 37.4   \\
&\textsc{xlm}                              & 53.7     & 53.2   & 57.1  & 52.0 \\        
&\textsc{re-lm}                           & \textbf{54.8}     & 53.4  & \textbf{58.1}  & \textbf{52.9}     \\ \midrule

\multirow{3}{*}{200K} &random   & 51.3     & 51.2   & 55.6 & 51.4   \\
&\textsc{xlm}            & 55.0      & 55.5  & 60.9 & 55.1 \\       
&\textsc{re-lm}         & 55.2      & 55.3  & 61.1 & 54.8  \\ 
\bottomrule

\end{tabular}
\caption{\textsc{bleu} scores on the dev set using increasing amounts of parallel data. We show in bold the models that achieve at least +$1$ \textsc{bleu} compared to \textsc{xlm}. }
\enote{DS}{Are the scores on dev?} \enote{AC}{yes (i think). is that a problem?}
\label{table:parallel}
\end{table}

\subsection{Analysis}
\label{ssec:analysis}
We experiment with different fine-tuning schemes and show results in Table \ref{table:ablation}. Then, we vary the number of BPE merges used to split the \textsc{lmr} language using the vocabulary extension method and also show experiments where this method is not used at all. The results are presented in Table \ref{table:vocext}. 

\noindent \textbf{\textsc{re-lm}.} In Table \ref{table:ablation}, we compare fine-tuning an \textsc{lm} \textit{only} on the \textsc{lmr} language to fine-tuning it on \textit{both} the \textsc{hmr} and \textsc{lmr} language (rows $1$ and $2$). Fine-tuning only on the \textsc{lmr} language provides worse \textsc{bleu} scores because of catastrophic forgetting. The negative effect is clear for Mk and Sq, where fine-tuning only on the \textsc{lmr} results in worse \textsc{bleu} scores than random initialization, shown in Table \ref{table:results}. 
For De, the effect is smaller, perhaps because En and De are very similar languages.

\begin{table*}[ht]
\centering
\small

\begin{tabular}{clrrrrrr|rrrr}
\toprule
  \multicolumn{2}{c}{\textsc{hmr}-\textsc{lmr} language pair}   &  \multicolumn{2}{c}{\textbf{En-De}}  & \multicolumn{2}{c}{\textbf{En-De}}   & \multicolumn{2}{c|}{\textbf{\textbf{En-De}}} & \multicolumn{2}{c}{\textbf{En-Mk}}           & \multicolumn{2}{c}{\textbf{En-Sq}}     \\
  \multicolumn{2}{c}{size of \textsc{lmr} language}   & \multicolumn{2}{c}{\textbf{0.05M}}  & \multicolumn{2}{c}{\textbf{0.5M}}   & \multicolumn{2}{c|}{\textbf{1M}} & \multicolumn{2}{c}{\textbf{2.4M}}           & \multicolumn{2}{c}{\textbf{4M}} \\
    &  &  $\leftarrow$  &  $\rightarrow$      &   $\leftarrow$  &  $\rightarrow$    &   $\leftarrow$  &  $\rightarrow$       &    $\leftarrow$  &  $\rightarrow$   &   $\leftarrow$  &  $\rightarrow$    \\  \midrule
 \multirow{4}{*}{\textsc{lm}} & ft on \textsc{lmr} &  9.4 & 7.3 &  20.4 & 16.8 & 20.6 & 17.8 & 2.7    & 2.4       & 4.7 & 4.7 \\ 
  & ft on \textsc{lmr \& hmr} (\textbf{\textsc{re-lm}})   & \textbf{10.7} & \textbf{7.5}  &\textbf{22.6}   & \textbf{19.0}   & \textbf{24.3} &\textbf{21.9}  & \textbf{22.0}   & \textbf{21.1}    &  27.6    & 28.1 \\
& + adapters ft on \textsc{lmr} (\textbf{adapter} \textbf{\textsc{re-lm}})     & 9.8 &  \textbf{7.5} & 21.3  & 18.3 & 23.7 & 20.0  & 21.6    &  19.0 &  \textbf{30.2}  & \textbf{29.4}    \\
& + adapters ft on \textsc{lmr \& hmr}   & 9.2 & 7.1 & 20.6 & 18.0 & 23.4 & 19.9 & 21.6    & 20.3    & 24.6 & 25.5 \\
\bottomrule

\end{tabular}
\caption{Comparison of \textsc{unmt} \textsc{bleu} scores obtained using different fine-tuning schemes of the pretrained monolingual \textsc{lm}. \textit{\textit{\small{LM}}} refers to the pretrained \textsc{lm} (on  \textsc{hmr} data), while \textit{ft} refers to fine-tuning.}

\label{table:ablation}
\end{table*}
\noindent \textbf{Adapters.} We insert adapters to the pretrained \textsc{lm} and fine-tune only the adapter and embedding layer. We use the fine-tuned \textsc{lm} to initialize a \textsc{unmt} system. Adapters are used for both translation directions during \textsc{unmt} training. Results are presented in Table \ref{table:ablation}. 
Fine-tuning the \textsc{lm} only on the \textsc{lmr} language yields at least $+3.9$ \textsc{bleu} for En-Sq compared to fine-tuning on both (rows $3$, $4$). 
En and Sq are not similar languages and their embeddings also differ. Thus, fine-tuning on both is not helpful. By contrast, fine-tuning only on Sq preserves the pretrained model's knowledge, while adapters are trained to encode Sq. 
For En-De and En-Mk, both approaches provide similar results. En and Mk do not share an alphabet, so their embeddings do not overlap and both fine-tuning methods are equally effective. In En-De, fine-tuning only on De is marginally better than fine-tuning on both. 
We highlight that adapters allow parameter-efficient fine-tuning. Adapter \textsc{re-lm} reaches almost the same results as \textsc{re-lm}, using just a fraction of the \textsc{re-lm} parameters while fine-tuning. 
Details can be found in the appendix.

\begin{table}[ht]
\centering
\small
\begin{tabular}{lrrrrrr}
\toprule
 &  \multicolumn{2}{c}{\textbf{\textbf{En-De}}} & \multicolumn{2}{c}{\textbf{En-Mk}}       & \multicolumn{2}{c}{\textbf{En-Sq}}     \\
  $\text{BPE}_{joint}$   &\multicolumn{2}{c}{\textbf{0.5M}}   & \multicolumn{2}{c}{\textbf{2.4M}}      & \multicolumn{2}{c}{\textbf{4M}} \\
 merges &  $\rightarrow$   &   $\leftarrow$  &  $\rightarrow$  &    $\leftarrow$ &   $\rightarrow$    &    $\leftarrow$    \\ \midrule
   - & 8.1  & 8.0 & 6.1 & 6.4  & 7.2 & 7.6  \\  \midrule
\textbf{8K} & 8.3 & 10.2  & 14.3  & 17.3 & 18.1 & 16.4 \\
 \textbf{16K} & 8.7  & 14.6  & 14.9    & 20.2 & 27.1 & 25.5 \\
  \textbf{32K}  &22.6  & 19.0  & 22.0   & 21.1    &  27.6    & 28.1 \\
\bottomrule

\end{tabular}
\caption{\textsc{unmt} \textsc{bleu} scores obtained with \textsc{re-lm}, \textbf{with} (rows $2$-$4$) and \textbf{without} (row $1$) extending the vocabulary of the pretrained \textsc{lm} ($V_{\text{\textit{HMR}}}$). When extending the vocabulary, we vary the number of $\text{BPE}_{joint}$ merges used to split the \textsc{lmr} data. We note that $32$K BPEs are used to split the \textsc{hmr} data ($\text{BPE}_{\text{\textit{HMR}}}$).}

\label{table:vocext}
\end{table}

\begin{table}[ht]
\centering
\small
\begin{tabular}{lrrr}
\toprule
 $\text{BPE}_{joint}$  & \multicolumn{3}{c}{new vocabulary items}  \\ 
merges  & Mk & Sq & De \\  \midrule
\textbf{8K} &  5K& 5K &  0.6K \\
\textbf{16K} &  10K  &  10K & 2K  \\ 
\textbf{32K}  & 19K & 20K &  19K   \\
\bottomrule
\end{tabular}

\caption{Statistics of the vocabulary extension method. We split the \textsc{lmr} corpus using 8K, 16K, or 32K BPE merges and report the number of new vocabulary items.}

\label{table:stats}
\end{table}

\noindent \textbf{Vocabulary Extension.} In order to use \textsc{re-lm}, we extend the vocabulary of each language, as described in \S \ref{ssec:voctrick}. The intuition is that, since the pretrained monolingual \textsc{lm} uses BPEs learned exclusively on the \textsc{hmr} language, these BPEs would not split the \textsc{lmr} corpus in a meaningful way. We conduct experiments to clarify the contribution of the vocabulary extension, presented in Table \ref{table:vocext}.  In Table \ref{table:stats}, we present the amount of vocabulary items added for each of our experimental setups. 

Without vocabulary extension, the results are poor. This is expected, as in the case of Mk for example, the \textsc{hmr} language (En) uses Latin alphabet, whereas Mk uses Cyrillic. If the vocabulary of Mk is not taken into account, the \textsc{unmt} model cannot provide accurate results. The same applies for Sq and De. We hypothesize that, even though these languages use Latin script, a lot of their words do not appear in En, therefore extending the initial vocabulary to include them is crucial. 
Using vocabulary extension, we experiment with learning $8$K, $16$K or $32$K BPEs on the joint corpus. We then use them to split the \textsc{lmr} data. We observe in Table \ref{table:vocext} that even using only $8$K BPEs, there is a large improvement in Mk and Sq (more than +$8$ BLEU). For En-De, 
the improvement is negligible. This might be the case because, as Table \ref{table:stats} shows, using $8$K merges, only $600$ items are added to the initial vocabulary, which are not sufficient for representing De language. This setup for En-De is in fact very similar to not employing vocabulary extension. We notice that adding more vocabulary items (using more BPE merge operations) is helpful for all language pairs, providing improved BLEU scores.

\section{Conclusions}
Training competitive unsupervised \textsc{nmt} models for \textsc{hmr}-\textsc{lmr} scenarios 
is important for many real low-resource languages. 
We proposed \textsc{re-lm}, a novel approach that fine-tunes 
a high-resource \textsc{lm} on a low-resource language and initializes an \textsc{nmt} model. \textsc{re-lm} outperformed a strong baseline in \textsc{unmt}, while also improving translations on a low-resource supervised setup. In future work, we will apply our method to languages with corpora from diverse domains and also to other languages.

\section*{Acknowledgments}

This project has received funding from the European Research Council under the European Union’s Horizon $2020$ research and innovation program  (grant agreement 
\#$640550$). This work was also supported by DFG (grant FR $2829$/$4$-$1$). We thank Katerina Margatina and Giorgos Vernikos for their valuable comments and help with the first draft of this paper.

\bibliographystyle{acl_natbib}
\bibliography{emnlp2020}

\appendix
\clearpage
\appendix
\section{Appendix}

\subsection{Vocabulary Extension}
\begin{figure}[ht]
	\centering
	\includegraphics[width=1\columnwidth, page=1]{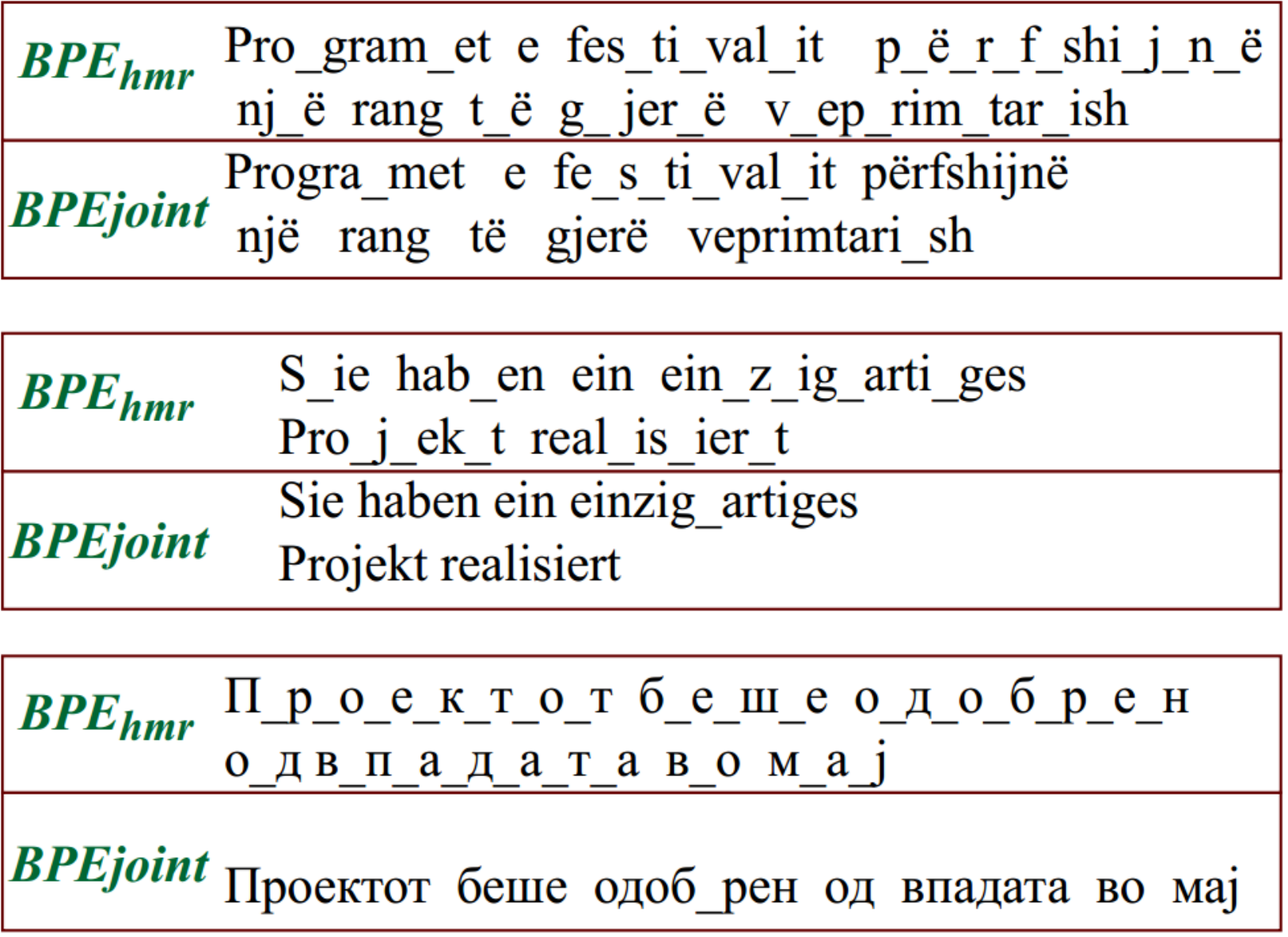}
	\caption{Segmentation of Sq, De and Mk using $\text{BPE}_{\text{\textit{HMR}}}$ or $\text{BPE}_{joint}$ tokens. Using $\text{BPE}_{\text{\textit{HMR}}}$ tokens results in heavily split words.}
	\label{fig:segmore}
\end{figure}

\noindent We provide more examples of different segmentations of Sq, De and Mk using either the $\text{BPE}_{\text{\textit{HMR}}}$ or the $\text{BPE}_{joint}$ tokens in Figure \ref{fig:segmore}. We observe that, as expected, the Mk sentence is split to the character level, as it uses a different alphabet (Cyrillic) than the one that the $\text{BPE}_{\text{\textit{HMR}}}$ tokens were learned on (Latin).

\subsection{Datasets} 
 We report that we remove sentences longer than $100$ words after BPE splitting. We split the data using the fastBPE codebase\footnote{https://github.com/glample/fastBPE}. 

\subsection{Model Configuration}
 We tie the embedding and output (projection) layers of both \textsc{lm} and \textsc{nmt} models \cite{press-wolf-2017-using}. We use a dropout rate of $0.1$ and  GELU activations \cite{hendrycks2016bridging}. We use the default parameters of \citet{lample2019cross} in order to train our models unless otherwise specified. 
We do not tune the hyperparameters. The code was built with PyTorch \cite{paszke2019pytorch} on top of the \textsc{xlm} implementation\footnote{https://github.com/facebookresearch/XLM/}. This code was used for \textsc{lm} pretraining, \textsc{lm} fine-tuning, \textsc{unmt} training, and \textsc{nmt} training. 

\vspace{5pt}
\noindent \textbf{LM configuration and training details.} 
 \textsc{re-lm} approach pretrains a \textbf{monolingual} language model whereas the \textsc{xlm} approach pretrains a \textbf{bilingual} language model. We obtain a checkpoint every $200$K sentences processed by the model. We train each \textsc{lm} using as criterion the validation perplexity on the \textsc{lmr} language, with a patience of $10$. 

The training details of the two \textit{pretraining} methods are presented here:

\begin{itemize}
    \item  The monolingual \textsc{lm} pretraining required $1$ week, $8$ GPUs and had $137$M parameters.
    \item The \textsc{xlm} pretraining required $1$ week, in $8$ GPUs. The total number of trainable parameters is $138$M.
\end{itemize}

\noindent Our approach also requires an \textit{LM fine-tuning} step. The runtimes, parameters and GPU details are shown in Table \ref{table:finetuning} under \textsc{re-lm} \textit{ft} column. The runtimes mentioned refer to the En-Mk language pair. We note that the \textit{LM fine-tuning} step is a lot faster than performing \textit{XLM pretraining} for each language pair (note that pretraining ran on $8$ GPUs, whereas fine-tuning on a single GPU).

\vspace{5pt}
\noindent \textbf{NMT configuration and training details.} 
The parameters and runtimes of the \textsc{unmt} models we used are shown in Table \ref{table:finetuning} under  \textsc{unmt} columns.
Likewise,
the details of supervised \textsc{nmt} models are shown under sup \textsc{nmt} columns. We get a checkpoint every $50$K sentences processed by the model.  
Regarding the adapter \textsc{re-lm} training procedure, we note that, different from  \citet{Houlsby2019ParameterEfficientTL,bapna-firat-2019-simple}, we also freeze the layer normalization \cite{ba2016layer} parameters, without introducing new ones.

\subsection{Validation Scores of Results}

In Tables \ref{table:results1_dev} and \ref{table:results2_dev} we show the dev scores of the main results of our proposed approach (\textsc{re-lm}) compared to the baselines. These Tables extend Table \ref{table:results} of the main paper. 

In Tables \ref{table:ablation1_dev} and \ref{table:ablation2_dev}, we show the dev scores of the extra fine-tuning experiments we did for the analysis. The Tables correspond to Table \ref{table:ablation} of the main paper. 

We note that the dev scores are obtained using greedy decoding, while the test scores are obtained with beam search of size $5$. We clarify that we train each \textsc{nmt} model using as training criterion the validation \textsc{bleu} score of the \textsc{lmr}$\rightarrow$\textsc{hmr} direction, with a patience of $10$. We specifically use \texttt{multi-bleu.perl} script from Moses. 

\begin{table*}[t]
\centering
\small
\begin{tabular}{lcc|ccc|cc|cc}
\toprule 
& \multicolumn{2}{c|}{\textsc{xlm}}   & \multicolumn{3}{c|}{\textsc{re-lm}}        & \multicolumn{2}{c|}{adapter \textsc{re-lm}} & \multicolumn{2}{c}{random} \\ 
     & \textsc{unmt} & sup \textsc{nmt} &  ft & \textsc{unmt} & sup \textsc{nmt} & ft      & \textsc{unmt}      & \textsc{unmt}      & sup \textsc{nmt}      \\ \midrule 
params    & 223M      & 223M    & 156M        & 258M      & 258M    & 88M              & 270M           & 258M            & 258M          \\
runtime     & 48h       & 10h          & 60h         & 72h       & 10h     & 44h              & 20h            & 18h             & 15h           \\\bottomrule 

\end{tabular}
\caption{Parameters and training runtimes used for each experiment. We note that each of the experiments ran on a single GPU. \textit{ft} refers to the fine-tuning of the pretrained monolingual \textsc{lm}. Adapter \textsc{re-lm} refers to the addition of adapters to the \textsc{lm} and the \textsc{unmt} model. }
\label{table:finetuning}
\end{table*}

\begin{table*}[h]
\centering
\small
\begin{tabular}{lrrrr|rrrr|rrrr}
\toprule
 languages  & \multicolumn{12}{c}{\textbf{En-De}}    \\  
size of \textsc{lmr}   &  \multicolumn{4}{c}{\textbf{0.05M}}  & \multicolumn{4}{c}{\textbf{0.5M}}   & \multicolumn{4}{c}{\textbf{1M}}  \\ 
&   \multicolumn{2}{c}{$\leftarrow$}   &  \multicolumn{2}{c}{$\rightarrow$} &  \multicolumn{2}{c}{$\leftarrow$}  & \multicolumn{2}{c}{$\rightarrow$} &    \multicolumn{2}{c}{$\leftarrow$}  &  \multicolumn{2}{c}{$\rightarrow$}\\  \cmidrule{2-13} 
& dev  & test & dev     & test  & dev   & test & dev     & test  &  dev     & test & dev     & test   \\ \midrule
random          & 3.2 & 3.9 & 4.1 & 4.9  & 2.5  &   3.4  & 2.3 & 2.6 & 3.7    & 4.2 & 3.5  & 4.1    \\
\textsc{xlm}    & 5.6 & 8.1  & 4.8 & 6.4 & 14.5 & 19.8   & 12.0 & 16.0 & 17.4   & 21.7& 14.6 &  18.1   \\
\textsc{re-lm} & 7.4 & 10.7 & 4.1 & 7.5 & 16.2  &  22.6 & 13.8 &  19.0  & 17.8 & 24.3 & 16.3  &  21.9  \\\bottomrule
\end{tabular}
\caption{Unsupervised \textsc{nmt} results with dev scores. The first column indicates the pretraining method used. \textit{Random} refers to random initialization, while \textsc{xlm} refers to the method of \citet{lample2019cross} and  \textsc{re-lm} to our proposed approach. }
\label{table:results1_dev}
\end{table*}

\begin{table*}[ht]
\centering
\small
\begin{tabular}{lrrrr|rrrr}
\toprule 
size of \textsc{lmr}   & \multicolumn{4}{c|}{\textbf{2.4M}}           & \multicolumn{4}{c}{\textbf{4M}}   \\
    &    \multicolumn{2}{c}{\textbf{Mk$\rightarrow$En}} &  \multicolumn{2}{c|}{\textbf{En$\rightarrow$Mk}}  &   \multicolumn{2}{c}{\textbf{Sq$\rightarrow$En}}   &  \multicolumn{2}{c}{\textbf{En$\rightarrow$Sq}} \\ \cmidrule{2-9} 
             & dev     & test & dev     & test  &      dev     & test & dev     & test   \\  \midrule
random          & 3.1 & 3.5 & 3.0 & 3.0  & 5.8  & 6.6 & 5.6 & 5.6      \\
\textsc{xlm}    & 11.8  & 12.2 & 12.6  & 12.8  & 15.5 & 16.3 & 17.3 & 18.8  \\
\textsc{re-lm} & 22.0 & 22.0  & 19.5 & 21.1  & 27.2 &  27.6  & 27.6 & 28.1 \\ \bottomrule
\end{tabular}

\caption{Unsupervised \textsc{nmt} \textsc{bleu} scores with corresponding dev scores for En-Mk, En-Sq. }
\label{table:results2_dev}
\end{table*}

\begin{table*}[ht]
\centering
\small
\begin{tabular}{clrrrr|rrrr|rrrr}
\toprule
\multicolumn{2}{c}{languages}  & \multicolumn{12}{c}{\textbf{En-De}}    \\  
\multicolumn{2}{c}{size of \textsc{lmr}}    &  \multicolumn{4}{c}{\textbf{0.05M}}  & \multicolumn{4}{c}{\textbf{0.5M}}   & \multicolumn{4}{c}{\textbf{1M}}  \\ 
& &   \multicolumn{2}{c}{$\leftarrow$}   &  \multicolumn{2}{c}{$\rightarrow$} &  \multicolumn{2}{c}{$\leftarrow$}  & \multicolumn{2}{c}{$\rightarrow$} &    \multicolumn{2}{c}{$\leftarrow$}  &  \multicolumn{2}{c}{$\rightarrow$}\\  \cmidrule{3-14} 
& & dev  & test & dev     & test  & dev   & test & dev     & test  &  dev     & test & dev     & test   \\ \midrule
 \multirow{4}{*}{\textsc{lm}} & ft \textsc{lmr}      & 6.8 & 9.4 & 5.2 & 7.3 & 15.1 & 20.4 & 12.9 &16.8  & 15.3 & 20.6 & 13.3 & 17.8   \\
& ft both (\textbf{\textsc{re-lm}})& 7.4 & 10.7 & 4.1 & 7.5 & 16.2  &  22.6 & 13.8 &  19.0  & 17.8 & 24.3 & 16.3  &  21.9  \\
&  \textbf{+ adapter \textsc{re-lm}}  & 6.8 & 9.8 & 4.8 & 7.5 & 15.1 & 21.3 & 13.4 & 18.3 & 16.9 & 23.7 & 15.2 & 20.0 \\
&  + adapters ft both & 6.7 & 9.2 & 4.1 & 7.1 & 14.8 & 20.6 & 13.0 & 18.0 & 17.1 & 23.4 & 15.0 & 19.9  \\
\bottomrule
\end{tabular}
\caption{Comparison of \textsc{unmt} \textsc{bleu} scores obtained using different fine-tuning schemes of the pretrained monolingual \textsc{lm} with corresponding dev scores for En-De. \textit{\textit{\small{LM}}} refers to the pretrained \textsc{lm}, trained on  \textsc{hmr} data, while \textit{ft} refers to fine-tuning. \textit{ft both} means fine-tuning on the \textsc{lmr} and the \textsc{hmr} language.} 
\label{table:ablation1_dev}
\end{table*}

\begin{table*}[ht]
\centering
\small
\begin{tabular}{clrrrr|rrrr}
\toprule 
\multicolumn{2}{c}{size of \textsc{lmr}}   & \multicolumn{4}{c|}{\textbf{2.4M}}           & \multicolumn{4}{c}{\textbf{4M}}   \\
&    &    \multicolumn{2}{c}{\textbf{Mk$\rightarrow$En}} &  \multicolumn{2}{c|}{\textbf{En$\rightarrow$Mk}}  &   \multicolumn{2}{c}{\textbf{Sq$\rightarrow$En}}   &  \multicolumn{2}{c}{\textbf{En$\rightarrow$Sq}} \\ \cmidrule{3-10} 
     &        & dev     & test & dev     & test  &      dev     & test & dev     & test   \\  \midrule
\multirow{4}{*}{\textsc{lm}} & ft \textsc{lmr}           & 2.6 & 2.7 & 2.3 & 2.4 & 4.4 & 4.7 & 4.2 & 4.7     \\
& ft both (\textbf{\textsc{re-lm}})  & 22.0 & 22.0  & 19.5 & 21.1  & 27.2 &  27.6  & 27.6 & 28.1  \\
& + \textbf{adapter \textsc{re-lm}} & 21.4 & 21.6 & 20.0 & 19.0 & 29.8 & 30.2 & 29.3 & 29.4 \\
& + adapters ft both  & 22.7 & 21.6 & 22.2 & 20.3 & 24.4 & 24.6 & 25.4 & 25.5 \\  \bottomrule

\end{tabular}

\caption{Comparison of \textsc{unmt} \textsc{bleu} scores obtained using different fine-tuning schemes of the pretrained monolingual \textsc{lm} with corresponding dev scores for En-Mk and En-Sq. \textit{\textit{\small{LM}}} refers to the pretrained \textsc{lm}, trained on  \textsc{hmr} data, while \textit{ft} refers to fine-tuning. \textit{ft both} means fine-tuning on the \textsc{lmr} and the \textsc{hmr} language. }

\label{table:ablation2_dev}
\end{table*}

\end{document}